\documentclass{article}

\usepackage[square,sort,comma,numbers]{natbib}

\usepackage[final]{nips_2016}

\usepackage[utf8]{inputenc} 
\usepackage[T1]{fontenc}    
\usepackage{url}            
\usepackage{booktabs}       
\usepackage{amsfonts}       
\usepackage{nicefrac}       
\usepackage{microtype}      
\usepackage{amsmath}
\usepackage{amssymb}
\usepackage{graphicx}
\usepackage{subfig}
\usepackage{color}

\title{Associative Adversarial Networks}
\author{
  Tarik Arici, Asli Celikyilmaz \\
  Microsoft AI and Research\\
  Redmond, WA \\
  \texttt{\{taaric, aslicel\}@microsoft.com } \\
}

\begin{document}

\maketitle

\begin{abstract}
We propose a higher-level associative memory for learning adversarial networks. Generative adversarial network (GAN) framework has a discriminator and a generator network. The generator (\textit{G}) maps white noise ($z$) to data samples while the discriminator (\textit{D}) maps data samples to a single scalar. To do so, G learns how to map from high-level representation space to data space, and D learns to do the opposite. We argue that higher-level representation spaces need not necessarily follow a uniform probability distribution. In this work, we use Restricted Boltzmann Machines (RBMs) as a higher-level associative memory and learn the probability distribution for the high-level features generated by \textit{D}. The associative memory samples its underlying probability distribution and \textit{G} learns how to map these samples to data space. The proposed associative adversarial networks (AANs) are generative models in the higher-levels of the learning, and use adversarial non-stochastic models D and G for learning the mapping between data and higher-level representation spaces. Experiments show the potential of the proposed networks.
\end{abstract}

\section{Introduction}
The generative adversarial network (GAN) framework \cite{Goodfellow2014} is a relatively new type of framework that introduces a generator \textit{G} and a discriminator \textit{D}, both of which are often chosen to be a type of multilayer perceptron (MLP) network, and are trained in an adversarial manner. \textit{D} is trained to label training examples as true, and outputs of \textit{G} as false, while \textit{G} is trained to maximize the probability of \textit{D}'s classification errors.

Recent works have reported that models trained using the GAN framework (especially for the image generation task) can generate excellent samples \cite{RadfordMC15}. Nevertheless, it is widely reported that GAN models are difficult to train and different techniques are proposed (which are summarized in \cite{Salimans2016,OPENAI2016,Oriol2016}).
Most of these work try to deal with the difficulty of optimizing the \textit{G} and \textit{D} simultaneously in a synchronized manner and their joint convergence\cite{Goodfellow2015}. \cite{Salimans2016} proposes to train the generator by a new objective that guides \textit{G} to match the statistics of features on an intermediate layer of the discriminator. This helps to avoid over-training of the current discriminator and discourages learning mappings that are not useful in generating realistic data. Another major difficulty in training GANs is the tendency for \textit{G} to collapse by converging to a parameter setting which maps all $z$'s to the same data. \cite{Salimans2016} uses minibatch discrimination so that \textit{D} can generate different gradients for data samples and continue guiding \textit{G} via backpropagation and force \textit{G} to diversify its outputs and avoid collapsing.

In this paper, we argue that using \textit{noise} for the generator contributes to the difficulty of training GANs. The generator is assigned the task of learning the mapping from the input signal $z$, which is a uniformly distributed noise to data space. In \cite{Bengio2013}, a strong evidence is presented for better disentangling of the underlying factors of variation in data by higher-level representation spaces, and this might explain why a uniformly distributed random noise as input to \textit{G} works. However, learning the mapping from a flat representation space to data space is difficult. This can be explained with an example from human-face image generation task. A completely disentangled representation space for human faces would likely consist of features including pose, illumination, gender, mood, facial expressions, 3-D model of face etc. The generator's task is then to find good mappings from this flat representation space to face images, which requires significant training. It is difficult for the generator network to achieve this goal and this might be one of the reasons why generator collapses during training. However, mapping a higher-level representation that does not assume complete disentanglement of factors of variation, such as mapping facial areas to a face image, can be expected to be an easier task. This higher-level representation-space features will be correlated (\emph{e.g.}, the existence of vertical symmetry in the facial parts or the location of distinct facial areas depending on the pose). By using an associative memory on $z$ and generating samples from this memory as an input to \textit{G}, our goal is to alleviate \textit{G}'s learning task. Also, mapping data samples to a lower dimensional but higher-level representation space by using \textit{D} and then learning an associative memory model causes the Markov chain (used in the model training) to mix faster - thanks to a more uniform distribution in this new space \cite{Bengio2013}.

\begin{figure}[t]
\begin{center}
\includegraphics[width=1.0\textwidth]{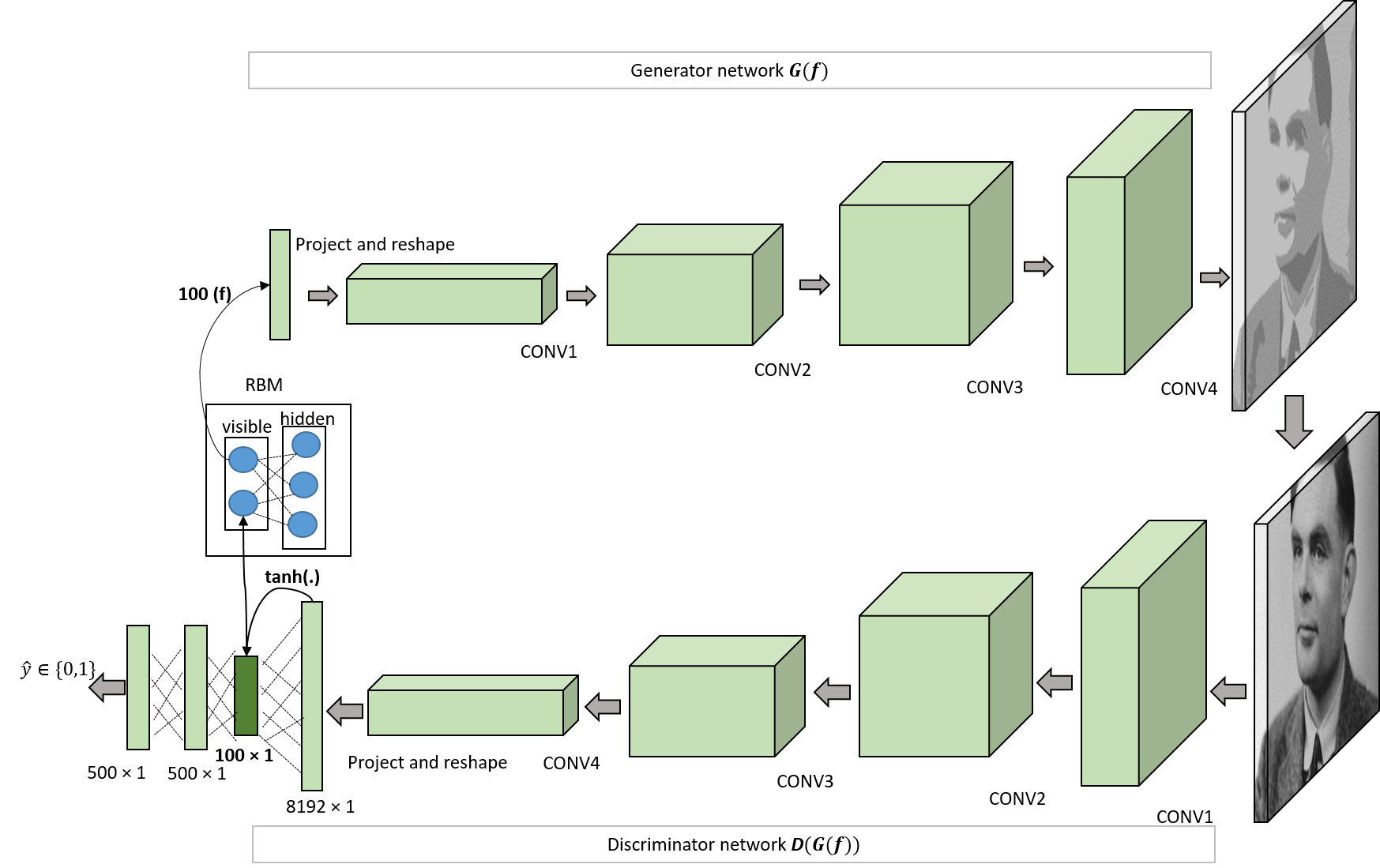}
\end{center}
\caption{AAN: The Associative Adversarial Network Model. An intermediate layer of the discriminator (depicted in dark green) is clamped as a visible layer to the RBM network (the associative memory), which is then sampled to generate inputs for the generator network (as opposed to noise sampling). This layer that is visible to the associative memory represents a feature space that can capture latent factors of variations in the data. The following layers in the discriminator mostly serve in the classification task embedded into GAN objective functions. However, there is no actual distinction between the network layers, and no such constraints are imposed. There are three separate networks, of which two are non-stochastic (D and G) models and one is a stochastic generative (RBM) model. All networks are jointly learned.}
\label{fig1}
\end{figure}

The associative adversarial networks (AANs) uses an associative memory that is located in between the generator and the discriminator networks, connecting the two networks together. We explore the special case that uses a Restricted Boltzmann Machine (RBM) as an associative memory, which learns a probability distribution over the features of an intermediate layer of the discriminator. Gibbs sampling is used to sample from this distribution, and the samples are fed to \textit{G}. Figure~\ref{fig1} depicts an example AAN, which we present in this work.

The main contributions of this work are the following: 
\begin{itemize}
  \setlength{\itemsep}{1pt}
  \setlength{\parskip}{0pt}
  \setlength{\parsep}{0pt}
\item We introduce a higher-level associative memory, a stochastic generative model that connects the two non-stochastic GAN models; the discriminator and the generator. 
\item Training on CelebA face image dataset, we show convincing evidence that the associative memory can learn a probabilistic model on the higher-level representation space of the discriminator. It can then produce samples to be used as an input to the generator thereby alleviating its learning task. 
\item We further evaluate the performance of the assoicative memory, RBM's performance, by investigating the held-out likelihood of the model for varying parameters.
\end{itemize}

\section{Related Work}
Neural network models that generate data have been getting more attention recently due to their ability to learn reusable feature representations from large unlabeled data sets as well as generate realistic data samples. The generative adversarial networks (GANs) \cite{Goodfellow2014}, variational autoencoders (VAE) \cite{Kingma2014}, generative maximum mean discrepeancy networks \cite{Dziugaite}, deep generative models\cite{Rezende2014}, generative moment matching networks \cite{Yujia2016}, etc., 
have shown that a deep generative network can learn a distribution over samples. Specifically, GANs have been a promising family of generative models in the computer vision field, because they can produce sharp images by applying stochastic gradient descent on well-defined loss functions. Some GAN extensions have looked at laplacian pyramid extensions \cite{Denton2015} showing higher quality images, a recurrent network
approach \cite{Gregor2015} and a de-convolution network approach \cite{Dosovitskiy2014} demonstrating reasonable success with generating natural images.
Several recent papers focus on improving the stability of training and the quality of generated GAN samples \cite{Denton2015,RadfordMC15,Im2016,Yoo2016}. 
Among recent ones (as recently summarized in \cite{Oriol2016})  the following pop out for stabilizing GAN training: \textit{balancing/freezing}: to prevent the generator or discriminator to outpace one another, \textit{minibatch discrimination}: to prevent the generator to collapse on to a single sample and enable back-propagation of gradients to improve weights, \textit{historical averaging} \cite{Salimans2016}, a technique common used in game theory that uses a historical average of parameters as a regularization term in optimization.

In this work we build up on these techniques. We use some of the DCGAN architectural suggestions proposed in Radford et al. \cite{RadfordMC15}, which uses strided convolutions in the initial layers of the discriminator and fractional-strided convolutions in the later layers of the generator, which will be discussed in the experiments section. In a way, our proposed technique resembles the learning algorithm for deep belief nets presented in \cite{Hinton06}, which utilizes a contrastive version of the wake-sleep algorithm proposed in \cite{Hinton2002}. Deep belief nets have a top-level undirected associative memory together with directed generative models in the lower levels. To generate samples from a model, Gibbs sampling is used in the top-level associative memory and a sample from this memory is passed down stochastically through the directed generative connections in the lower layers. The model is learned by performing a contrastive version of the wake-sleep algorithm after a greedy layerwise initialization process. In the wake phase, a stochastic "up-pass" starts and generative weights between two adjacent layers are updated locally to maximize the likelihood of upper layer samples generating lower level samples. In the sleep phase, a stochastic "down-pass" starts and recognition weights between two adjacent layers are updated similarly. Our proposed method uses D for the up-pass and G for the down-pass.

\section{Generative Adversarial Networks (GAN)}
GANs \cite{Goodfellow2014} are a class of generative models that pose the training process as a game between a generator network ($G$) and a discriminator network ($D$) both of which are non-stochastic models. The generator network, $G(z;\theta^{(G)})$, is typically chosen as a feed forward or convolutional neural network depending on the task. It produces samples, by transforming vectors of noise $z$ as $x=G(z;\theta^{(G)})$. The discriminator $D$, is trained by taking the samples from the generator, $p_{G}(x)$, as negative instances and from the real data $p_{data}(x)$ as positive instances and is trained to distinguish generated samples from the real (training) data. 

\textit{D} takes the output of \textit{G} and maps to a binary classification probability. Generator then tries to “trick” the discriminator by generating \textit{fake} samples. The learning framework is a two-player game and is cast as a minmax optimization of a differentiable objective and solved greedily by iteratively performing gradient descent steps to improve $G$ and $D$ and eventually reaching a Nash equilibrium \cite{wiki}. The GAN problem can be formulated as a zero-sum game (minimax) which has a \textit{distinguishability game value function}, $V(D,G)$ \cite{Goodfellow2015}:
\[
\min_{G}\max_{D}=\mathbb{E}_{x\sim p_{data}(x)}[\log D(x)]+\mathbb{E}_{z\sim p_{z}(z)}[\log (1 - D(G(z)))] 
\]

The first term in the cost function forces \textit{D} to label real data samples as one, while the second term forces \textit{D} to label fake data samples as zero. \textit{G} tries to fake \textit{D} into labeling its output as one so minimizes the given cost function, while \textit{D} tries to maximize it.

\section{Using RBMs as an Associative Memory for GANs} 
An RBM is an energy based model for unsupervised learning with an underlying undirected graph. It consists of two layers of binary stochastic units: one \textit{visible} layer $v$ representing the data and one \textit{hidden} layer $h$ for the latent variables. $w_{ij}$ determines the strength of the interaction between the hidden $h_j$ and visible $v_i$ units. An energy function between visible and hidden variables $E(v,h)$, the probability $P(v,h)$ and the partition function $Z$ is defined as below:
\[
E(v,h) = \frac{1}{2}\sum_i v_i^2-\sum_{i,j}v_ihjw_{ij}-\sum_i{v_i b_i}-\sum_j{h_j c_j} \ ;\ P(v,h)=\frac{e^{-E(v,h)}}{Z} \ ; \ Z=\sum_{x,y}e^{-E(x,y)}
\]
The probability of a data point (represented by the visible state $v$) is defined by marginalizing over the hidden variables $P(v)=\sum_hP(v,h)$.
The training data log-likelihood for one sample is:
\[
\phi=log P(v)=\phi^++\phi^- ; \ \ \phi^+=log\sum_he^{-E(v,h)} ; \ \ \ \phi^-=logZ=\sum_{x,y}e^{-E(x,y)}
\]
The gradient of the log-likelihood involves a positive and a negative term. The positive gradient is $\frac{\partial\phi^+}{\partial w_{ij}}=v_i \cdot P(h_j=1|v)$ but the negative gradient $\frac{\partial\phi^-}{\partial w_{ij}}$ is intractable and requires summation of all values of hidden and visible variables which grows exponentially. A common method is to approximate the expectation in the second term by generating samples. Contrastive Divergence (CD) algorithm \cite{Hinton2010} runs the Markov chain for a few steps after clamping the visible layer to data examples. Another technique is to use persistent chains without clamping the visible layer \cite{persistent}. In our experiments, we use the CD algorithm with two steps of alternating Gibbs sampling for learning the RBM model. 

\section{Associative Adversarial Networks (AAN)}
\label{aanm}
The discriminator model \textit{D} tries to discriminate between the real and fake data. In doing so, \textit{D} learns features that can explain factors of variation in data and uses these features to achieve its classification goal. On the other hand \textit{G} tries to map a low-dimensional input to data. Contrary to common approaches, instead of representing the \textit{G}'s input as a flat space, we think of it as an intermediate but higher-level representation corresponding to one of the intermediate layers of \textit{D}. We use an RBM to learn a distribution in this space. The samples generated via contrastive divergence for updating the RBM model is used an input to \textit{G}. Therefore we connect \textit{D} and \textit{G} through a high-level feature space.

Let $F(x)$ denote an intermediate layer activations of \textit{D}, while $C(y)$ denotes the operations in the remaining layers in \textit{D}. Then, $D(x)=C(F(x))$. The associative memory learns the distribution $p_f$ of $f=F(x)$. Similar to the regular GANs, AAN optimization becomes:

\begin{equation}
\min_{G} \max_{\hat{p}_{f}} \max_{D}=\mathbb{E}_{x\sim p_{data}(x)}[\log D(x)]+\mathbb{E}_{f\sim \hat{p}_{f}(f)}[\log (1 - D(G(f)))] + \mathbb{E}_{f\sim p_{f}(f)}[\log \hat{p}_{f}]\label{eq:cost}
\end{equation}
The second expectation is over the estimated $\hat{p}_{f}$ since G is inputted with samples from the associative memory. Minimizing the third expectation forces the associative memory to estimate the probability distribution function of $f$.

\section{Details of Associative Adversarial Training}
We trained the AANs on Large-scale CelebFaces Attributes (CelebA) and the MNIST dataset. Images in the dataset are linearly scaled to the [-1, 1] range. GAN models are trained with mini-batch sizes of 256. Similar to \cite{RadfordMC15}, we used an Adam optimizer. For the RBM learning, we used the same mini-batch size with a learning rate of 0.001 and stochastic gradient descent (SGD) with momentum. We picked a momentum rate of 0.8. A contrastive divergence with two steps is used to create the negative samples. We used leaky rectified linear activations (LeakyReLU) \cite{Maas2013} for \textit{D} with one exception to be discussed below and rectified linear activation \cite{Hinton2010relu} for \textit{G} as activation functions. Using an intermediate layer of \textit{D} with LeakyReLU activation as a visible layer to the RBM requires it to be a Gaussian RBM which does not work well in practice. Hence, we used \textit{tanh} activation for the layer that connects as a visible layer to a binary RBM. We chose binary variable states as (-1) and (+1) similar to spin states in an Ising model \cite{ising}. The expected value of a unit in the RBM is $p*1 + (1-p)*-1=2p-1$ where $p$ is the conditional probability of a variable to be (1), which is a sigmoid. Thus, the expected value $2p-1$ becomes a \textit{tanh} non-linearity. The negative samples created for RBM's contrastive divergence learning are used as inputs for the \textit{G}.

\begin{figure}[t]
\subfloat[$1000\times1000$ RBM for modeling $p_f$]{\label{fig:conva}\includegraphics[clip,width=0.50\columnwidth]{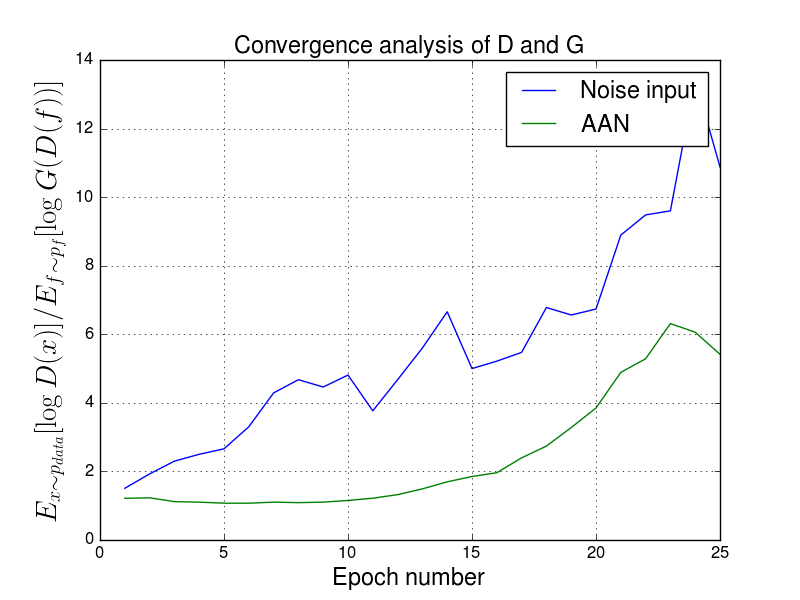}}
\subfloat[$100\times100$ RBM for modeling $p_f$]{\label{fig:convb}\includegraphics[clip,width=0.50\columnwidth]{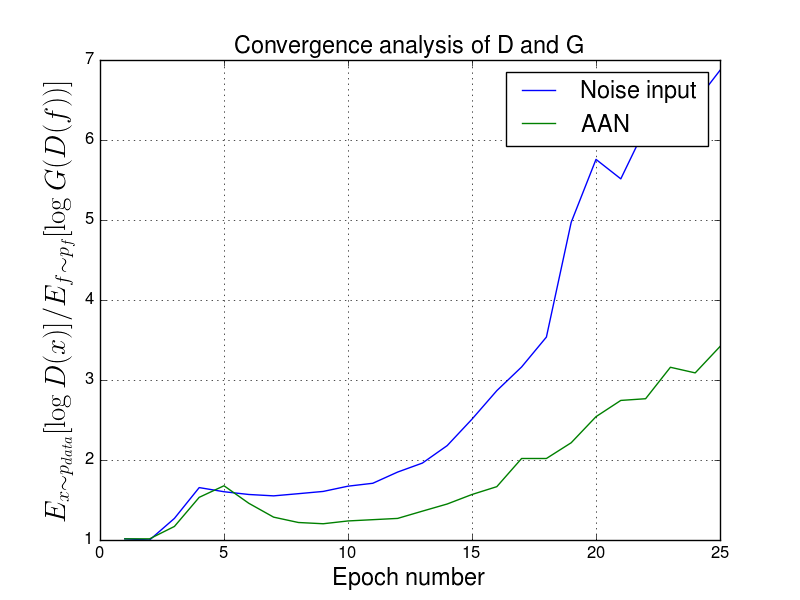}}
\caption{\label{fig:conv} AAN achieves a better adversarial training performance compared to feeding noise to G, since G more closely follows D in terms of maximizing $\log D()$ on real/generated images.}
\end{figure}

\begin{figure}[thp]
\subfloat[Using a $1000\times1000$ RBM]{\label{fig:a}\includegraphics[clip,width=\columnwidth]{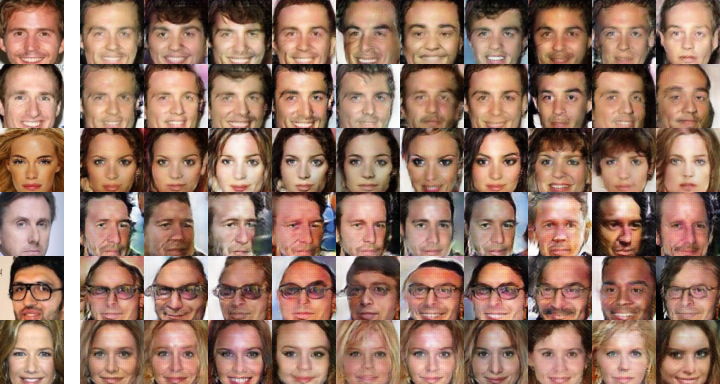}}\\
\subfloat[Using a $100\times100$ RBM]{\label{fig:b}\includegraphics[clip,width=\columnwidth]{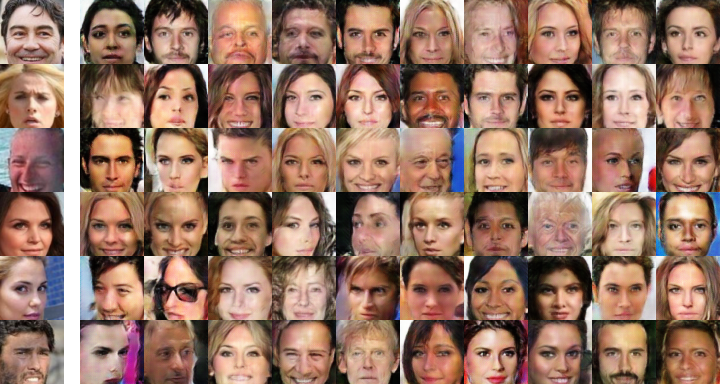}}
\caption{\label{fig:faces} Gibbs steps on faces - Face images generated by the model by running RBM's Markov chain with varying number of Gibbs sampling steps. Original images are in the leftmost column, while the other images are generated by increasing the number of Gibbs steps from one to ten with the last column corresponding to images generated by ten Gibbs steps. Note that images generated are not reconstructed versions of the original images since the discriminator and the generator networks are not designed to satisfy such a constraint. Images generated using a 1000-dimensional RBM change slowly with increasing number of Gibbs steps; facial features, expression and pose change smoothly. However, using a 100-dimensional RBM, generated images do not look correlated, implying that the Markov chain jumps from one mode of the distribution to another faster compared to a 1000-dimensional distribution. }

\end{figure}
\subsection{CelebA}
Discriminator for this dataset has four strided convolutional layers with depth 64, 128, 256, 512 as the first four layers. Stride size is set to match the filter width and height both of which is five. The last convolutional layer's outputs are reshaped into a one-dimensional representation $f$ which is fully connected to a layer that has the same dimension as the RBM's visible layer. After this layer there are two more layers of size 500. The two final layers can be thought as hidden layers of a classification network $C$ as defined in Section~\ref{aanm}, which has $f$ as input so that their cascaded application composes D (\textit{i.e.} $D(x)=C(F(x))$). $F$ defines a mapping to a feature space, and the RBM model learns the distribution of data set samples in this space.

Figure \ref{fig:faces} shows some face images generated using negative samples from the RBM generated by increasing the number of Gibbs steps from one to ten. RBM's Markov chain is initialized with $f$ created by real images from the data set which are given in the leftmost column. Figure \ref{fig:a} uses an RBM with 1000-dimensional visible and hidden layer units and Figure \ref{fig:b} uses an RBM with 100-dimensional visible and hidden layer units. With a $1000 \times 1000$ RBM, the generated images using the Gibbs samples change slowly; as more Gibbs steps are performed facial features, expressions change gradually. For example a pose or gender change does not happen quickly. This shows it requires many steps for the Gibbs sample to jump to other modes of the distribution. However, using a $100 \times 100$ RBM one sampling step is sufficient to change gender, race etc. as can be seen in Figure \ref{fig:b}. This shows that $f$ features are distributed more uniformly in the space, helping the Gibbs sample jump from one mode of the distribution to another. Compared to a 1000-D $f$, a 100-D $f$ has to be more efficient in packing the information to pass on to $C$ for the correct classification, so $F$ ends up learning a mapping that more uniformly distributes mapped dataset samples in this space.

\subsection{Convergence Analysis of the Generator and the Discriminator}
\textit{G} implicitly defines a probability distribution $p_G$ as the distribution of samples $G(f)$ it generates when $f \sim p_f$. The goal is to obtain a $p_G$ that is a good approximation to $p_{data}$. The minimax game that underlines the adversarial learning has a global optimum which is $p_G=p_{data}$ and $\log D(x)=1/2$. Therefore the value function given in \ref{eq:cost} is $\log 4$ at the optimal solution. As discussed in the introduction section, one of the difficulties with adversarial training is that \textit{D} improves its cost significantly faster than \textit{G}, and \textit{G} lags behind so much so that it can not receive good gradients back-propagating from \textit{D} and loses the game. For a good adversarial training, one would expect \textit{G} to lag behind at an acceptable level so that \textit{D} can guide \textit{G}. 

To analyze the convergence of adversarial models, we monitored $\mathbb{E}_{x\sim p_{data}(x)}[\log D(x)]$ and $\mathbb{E}_{f\sim \hat{p}_{f}(f)}[\log D(G(f))]$ during training. The first expectation measures the objective associated with real images and the second with generated images. If \textit{D} and \textit{G} both converge to the global optimal solution than both expectations should converge to $log 1/2$. However, in practice the first expectation becomes extremely close to zero while the second expectation diverges from zero. We analyzed the ratio $\mathbb{E}_{x\sim p_{data}(x)}[\log D(x)] / \mathbb{E}_{f\sim \hat{p}_{f}(f)}[\log D(G(f))]$ for AAN and GAN where noise $z$ is sampled uniformly to feed \textit{G}. Learning to map 1000-dimensional $f$ or $z$ to an image is more difficult as there is more learning associated with each dimension of \textit{G}'s input. Hence \textit{G} lags behind significantly. However, using a 100-dimensional input, \textit{G}'s learning task is less difficult. This can be seen in Figure \ref{fig:conv}. Similarly, AAN alleviates \textit{G}'s learning task by producing inputs from the manifold that $f$ lies in.

\section{Conclusion and Future Work}
We proposed associate adversarial networks for improving the training of generative adversarial networks (GAN). GANs are a promising class of generative models. However, previous works have reported several issues pertaining to its stability during training. In this work, we argue that inputting noise to the generator makes \textit{G}'s learning task difficult. Instead, we propose using an additional network, the RBM associative memory network, that connects the two networks of the GANs, the discriminator and generator. The associative memory networks can learn a probabilistic model using a higher level representation discovered by the discriminator and can be sampled to produce inputs for the generator network.

Although we empirically tested the efficacy of the proposed associative adversarial networks, we hope to develop a more rigorous theoretical understanding in future work. Inspecting equation (\ref{eq:cost}), one can see that it is possible for the associative memory to collapse to a degenerate probability distribution, similar to \textit{G}'s collapsing when it lags behind \textit{D} significantly. As a future work, we are planning to study entropy-maximizing regularizers for the associative memory.

Another future work would be to study probabilistic objectives for the generator, which can be obtained from the associative network. This is in agreement with the idea proposed in \cite{Salimans2016}, which suggest changing \textit{G}'s objective to match an intermediate discriminator layer's statistics.

\small
\bibliographystyle{unsrt}
\bibliography{Celik_Tur}

\end{document}